\documentclass[10pt,letterpaper,twocolumn]{article}

\usepackage[
  top=0.75in,
  bottom=0.80in,
  left=0.75in,
  right=0.75in
]{geometry}
\usepackage{amsmath,amssymb,amsthm}
\usepackage{newtxtext,newtxmath}
\usepackage{microtype}
\usepackage[T1]{fontenc}
\usepackage{chngcntr}
\usepackage{booktabs}
\usepackage{xcolor}
\usepackage[hidelinks]{hyperref}
\usepackage{multirow}

\usepackage[
  style=numeric,
  sorting=none,
  natbib=true
]{biblatex}
\newcommand{\KL}{\operatorname{KL}}

\newtheorem{theorem}{Theorem}
\newtheorem*{theorem*}{Theorem}

\newtheorem*{corollary*}{Corollary}
\newtheorem{proposition}{Proposition}

\theoremstyle{definition}

\newtheorem*{assumption*}{Assumption}

\theoremstyle{remark}

\title{PAC--Bayes Bounds on Quotient Parameter Spaces: Geometry-induced Implicit-Bias Priors}

\author{
Nicola Aladrah$^{1}$\thanks{\texttt{nicola.aladrah@phd.units.it}}
\quad
Fabio Anselmi$^{1,2}$\thanks{\texttt{fabio.anselmi@units.it}}
\\
$^{1}$Department of Mathematics, Informatics and Geoscience, University of Trieste, Via Valerio
\\12/1, 34127 Trieste, Italy\\
$^{2}$McGovern Institute, MIT, Main Street, Cambridge, MA 02139, USA
}

\date{}

\begin{document}

\maketitle

\begin{abstract}
Overparameterized models often have continuous parameter symmetries, so different parameters define the same predictor. We show that PAC--Bayesian analysis should be performed on the quotient predictor space: pushing a prior and posterior to the quotient preserves the empirical and population Gibbs risks while removing the nonnegative KL contribution caused solely by how the two distributions differ among parameterizations of the same predictor.

Quotienting alone does not determine which prior to use. We construct a canonical choice of one parameterization for each predictor and account for the geometric volume of its equivalent parameterizations. This transforms a neutral reference prior into a data-independent prior that reflects the model’s implicit bias. It approximates the ideal but inadmissible posterior-matched prior, which would minimize the KL term by depending on the training data. The resulting certificate is tighter exactly when this geometry-induced prior has smaller KL divergence from the learned quotient posterior than the neutral prior.

We test this prediction in Fourier regression with a Hadamard parameterization and in Query-Key attention, using ordinary SGD without an explicit regularizer. The implicit-bias prior reduces the mean quotient-space KL by \(40.69\%\) and the mean PAC--Bayes certificate by \(21.40\%\) in the Fourier-Hadamard experiment. The smaller, prior-scale-dependent improvement in Query-Key attention confirms the predicted conditional nature of the effect.

\end{abstract}
\section{Introduction}
Modern learning models usually operate in overparameterized regimes, where the number of trainable parameters exceeds the number of observations. Such models may contain large families of parameters that represent exactly the same predictor, as occurs, for example, in machine learning models with parameter symmetries~\cite{aladrah2026understanding, ziyin2025parameter}. To make this precise, let \(\mathcal X\) be the input space and \(\mathcal Y\) the output or label space. Let \(\mathcal{B}\) denote the parameter space, and let \(b\in\mathcal B\) be a parameter, such as the collection of weights of a model. Each \(b\) defines a predictor \(f_b:\mathcal X \to \mathcal Y\), which maps an input \(x \in \mathcal X\) to a prediction \(f_b(x)\in\mathcal Y\). The empirical risk is  the average loss of \(f_b\) on the observed training sample, whereas the population risk is its expected loss under the data-generating distribution. Let \(b\) and \(b'\) be two parameters related by such symmetry. If \(f_b = f_{b'}\) as functions on \(\mathcal X\), then they induce identical empirical and population risks for every loss that depends only on the predictor. This motivates our central goal: to formulate PAC--Bayesian complexity so that it reflects differences between predictors rather than differences between parameters that represent the same predictor. A PAC--Bayesian analysis performed directly on parameter space can nevertheless assign different prior and posterior masses to \(b\) and \(b'\), thereby charging their discrepancy in the Kullback--Leibler divergence. Its complexity term may therefore depend on representational degrees of freedom that have no effect on prediction or risk.

PAC--Bayesian theory bounds the population risk of a randomized predictor in terms of its empirical risk and a complexity penalty \(\KL(Q\Vert P)\), where \(P\) is a prior independent of the certification sample and \(Q\) is a data-dependent posterior over hypotheses~\cite{mcallester1998some,mcallester2013pac}. How should this framework be formulated for modern overparameterized models whose parameter spaces may contain multiple representations of the same predictor? When the parameter space itself is taken as the hypothesis space, the KL divergence can distinguish probability mass assigned to different parameters that define the same predictor. The resulting certificate consequently mixes predictor-level complexity with variation among equivalent parameter representatives. To address this problem, we apply the McAllester PAC--Bayes inequality~\cite{mcallester2003pac} on the quotient hypothesis space, where all the parameter representing the same predictor are identified. In this way, the symmetry structure of an overparameterized model becomes part of the specification of the PAC--Bayesian hypothesis space.

The symmetry considered here acts on parameter space rather than on the data. To make this concrete, consider the regression predictor, \(f_{a,c}(x)=(ac)x\), where the regression weight is \(w=ac\). The transformation, \((a,c)\mapsto (\lambda,a,\lambda^{-1}c)\), for \(\lambda>0\), leaves \(w\), and therefore the predictor unchanged. For a fixed \(w\), the set, \(\mathcal O_w=\{(\lambda a, \lambda^{-1} c):\lambda >0\}\), is \(\mathcal O_w\)-orbit or the symmetry orbit: its elements are different parameter values representing the same regression function. In general, for every group element \(g\) and parameter \(b\), \(f_{g\cdot b}=f_b\). The statistically relevant hypothesis is therefore the equivalence class of the parameters that implement one predictor. We represent this hypothesis space by the quotient of parameter space under the symmetry action, which collapses each symmetry orbit to a single quotient element. In the regression example, the whole orbit \(\mathcal O_w\) is represented by the single weight \(w\). Geometrically, this identification is illustrated by the hyperbolic symmetry orbits and the representative symmetry-breaking slice shown in Fig.~1 of the previous work~\cite{aladrah2026understanding}. This operation removes non-identifiable representational degrees of freedom. It neither imposes an invariance on the data distribution nor averages predictions from distinct predictors.

This quotient construction directly impacts the PAC–Bayesian complexity term. Let a parameter-space prior and posterior be pushed forward through the quotient map so that probability mass is assigned to the predictor classes represented by the parameters. Collapsing each symmetry orbit into a single quotient point does not change the predictor represented by that orbit. Hence, computing the empirical or population risk before this identification gives the same result as computing it afterward, and their posterior averages are unchanged. The Kullback–Leibler divergence behaves differently. On the parameter space, it measures how differently the prior and posterior distribute probability among the parameters that belongs to the same orbit. Once the orbit is collapsed to a single quotient point, this representational discrepancy disappears. The KL divergence on the quotient space therefore cannot be greater than the KL divergence on the parameter space. Applying the same PAC--Bayes inequality on the quotient thus preserves the risk terms while removing complexity that comes solely from differences between equivalent parameter representations.

This quotient construction is the first main contribution of work. It provides a general PAC--Bayesian reduction for predictor-preserving parameter symmetries: the complexity term cannot increase when equivalent parameterizations are identified, and it decreases whenever the posterior and prior distribute probability differently along symmetry orbits. This perspective follows in spirit the symmetry-based PAC--Bayesian analysis developed in~\cite{lyle2020benefits}. Here, however, the symmetry is formulated specifically on the parameter space: it relates different parameterizations of one predictor rather than transformations of the inputs or outputs. This parameter-space viewpoint connects naturally to the geometric mechanism of implicit bias prior induced by symmetry breaking, developed in~\cite{aladrah2026understanding}.

The quotient reduction, however, does not by itself select a preferred prior among the possible priors on the quotient space. This leads to our second contribution. The geometry of the symmetry provides a canonical way to represent each set of equivalent parameters by a single point. It also assigns a geometric weight to each predictor according to the volume of its redundant parameterizations. This weight defines an \textit{implicit-bias prior} on the quotient space, which defines a preference determined by the model architecture rather than by the data or by an added training penalty. This prior gives a tighter PAC--Bayes certificate than a neutral quotient prior only if it has smaller Kullback--Leibler divergence from the learned posterior. We evaluate this explicit posterior-dependent condition numerically. Thus, quotienting gives a general complexity reduction, while the additional benefit of the implicit-bias prior must be verified for the learned posterior.

Our results distinguish two effects. First, identifying the parameters that represent the same predictor leaves the risk terms unchanged and cannot increase the complexity term for any prior independent of the certification sample and any posterior. Second, choosing the implicit-bias prior does not necessarily improve the certificate. It helps only when this prior is closer to the chosen posterior, in KL divergence, than the neutral reference prior. We derive the exact condition for this comparison and test it in controlled numerical experiments.

\subsection{Related Work}
PAC--Bayesian theory gives high-probability bounds on the population risk of randomized predictors by comparing a data-dependent posterior with a prior chosen independently of the sample~\cite{mcallester1998some,mcallester2013pac}. Subsequent work has developed PAC--Bayes-kl inequalities, alternative loss-dependent bounds, and formulations for settings such as linear regression and Gaussian-process classification~\cite{seeger2002pac,catoni2007pac,alquier2016,shalaeva2020improved}. Our analysis takes a different starting point. Rather than modifying the PAC--Bayes inequality or introducing another loss-dependent bound, we examine how the hypothesis space and the prior should be defined when the parameterization contains multiple representations of the same predictor. We therefore use the symmetry structure of the parameter space to pass to a quotient hypothesis space and then use the associated geometric correction to construct a prior on that quotient. Our analysis uses a bounded certification loss and evaluates the posterior--prior KL divergence explicitly.

Parameter symmetries are a central source of non-identifiability in modern learning models. Continuous rescaling and change-of-basis symmetries can leave the represented predictor unchanged while modifying optimization trajectories, local geometry, and parameter-space descriptions of complexity~\cite{ziyin2025parameter,zhao2025symmetry,simsek2021geometry,zhao2024improving,godfrey2022symmetries}. In a complementary geometric analysis of stochastic learning, symmetry breaking yields a volume correction in the reduced distribution and thereby induces an effective preference among equivalent parameter representations that can be treated as the implicit bias of the model architecture~\cite{aladrah2026understanding}. We use this correction for a different purpose: to construct and compare quotient-space priors for PAC--Bayesian certification.

A related PAC--Bayesian literature studied symmetries of the learning problem, where a group acts on inputs and outputs and the hypothesis class is required to be invariant or equivariant. PAC--Bayesian generalization benefits have been established for compact group symmetries with invariant data distributions~\cite{lyle2020benefits}, and subsequently extended to non-compact groups and non-invariant data distributions~\cite{beck2025symmetries}. These works concern how transformed inputs and outputs are related by a predictor. Our setting is different: the symmetry acts only on the parameters and leaves the predictor unchanged. For Hadamard parameterization and other learning models that admit symmetries, the appropriate operation is therefore to identify parameter representatives of one predictor before evaluating the KL divergence, rather than to average over transformed inputs or impose predictor equivariance.

Quotient constructions and reparameterization are also common in learning models with non-identifiable parameters. The PAC--Bayesian approach is more specific: KL divergence is defined between probability measures on the selected hypothesis space. When equivalent parameters are treated as distinct elements of the parameter-space, the KL term can include posterior--prior mismatch within a predictor equivalence class. The quotient formulation makes this contribution explicit by preserving posterior-averaged risks under pushforward while contracting the KL term. It therefore separates predictor-level complexity from representational complexity.
\section{PAC--Bayesian bound on quotient parameter space}
Let \(\mathcal B^\circ\) denote the regular part of the parameter space, equipped
with its Borel \(\sigma\)-algebra. Each parameter \(b\in \mathcal B^\circ\) defines a measurable predictor \(f_b:\mathcal X\to\mathcal Y\), where \(\mathcal X\) is the input space and \(\mathcal Y\) is the output space. Let \(\mathcal D\) be the data-generating distribution on \(\mathcal X\times\mathcal Y\), and let \(S=(Z_1,\ldots,Z_n)\sim\mathcal D^n\) for \(Z_i=(X_i,Y_i)\), be a certification sample of size \(n\). We use a measurable loss \(\ell:\mathcal Y\times\mathcal Y\to[0,1]\). For \(b\in \mathcal B^\circ\), its population and empirical risks are
\begin{align}
R^b(b)
&:=
\mathbb E_{(X,Y)\sim\mathcal D}
\left[\ell\bigl(f_b(X),Y\bigr)\right],
\\
\widehat R_S^b(b)
&:=
\frac{1}{n}\sum_{i=1}^{n}
\ell\bigl(f_b(X_i),Y_i\bigr).
\end{align}

Let \(P^b\) be a prior probability measure on \(\mathcal B^\circ\), chosen independently of the certification sample \(S\), and let \(Q^b\) be a
posterior probability measure on \(\mathcal B^\circ\), which may depend on \(S\). For a confidence parameter \(\delta\in(0,1)\), the McAllester PAC--Bayes bound states that, with probability at least \(1-\delta\) over the draw of \(S\),
\begin{align}
\mathbb E_{b\sim Q^b}[R^b]
&\leq
\mathbb E_{b\sim Q^b}[\widehat R_S^b] + \operatorname{Comp}_{n,\delta}(\KL(Q^b\Vert P^b)).
\label{eq:mcallester-parameter}
\end{align}
where,
\begin{align}
    \operatorname{Comp}_{n,\delta}(\KL(Q^b\Vert P^b)) &= \nonumber\\ &\sqrt{ \frac{ \KL(Q^b\Vert P^b) +\log(1/\delta) +\log n +2}{2n-1}}
\end{align}
Here \(\KL(Q^b\Vert P^b)\) is the Kullback--Leibler divergence between the posterior and the prior~\cite{mcallester2003pac}. We denote the complexity term by \(\operatorname{Comp}_{n,\delta}(\KL(\cdot\Vert\cdot))\).

To introduce the notion of quotient, we present first the symmetry reduction. Let \(\mathcal G\) be a group acting measurably on \(\mathcal B^\circ\), and let \(U:=B^\circ/\mathcal G\) be the corresponding quotient space. Its elements are the
\(\mathcal G\)-orbits of the parameters. We assume that \(U\) is a measurable space and denote the quotient map by \(\phi:B^\circ\to U\), where \(\phi(b)=[b]\). Here \([b]\) is the orbit containing \(b\). The quotient prior and
posterior are the pushforwards \(P^u:=\phi_\#P^b\) and \(Q^u:=\phi_\#Q^b\). Thus, \(P^u\) and \(Q^u\) assign probability to symmetry orbits rather than
to individual parameter representatives. Since \(P^b\) is independent of \(S\), its pushforward \(P^u\) is also independent of \(S\) and is therefore
an appropriate PAC--Bayesian prior on the quotient space.

To justify replacing the parameter space by its quotient, we must check the two parts of the PAC--Bayesian certificate separately. First, we assume that identifying equivalent parameters does not change the empirical or population risk. Second, we must compare the KL divergence before and after the identification. The first point holds whenever the symmetry changes only the parameter representation and leaves the predictor unchanged.
\begin{assumption*}[Parameter-space symmetry]
\label{assumption:1}
For every \(g\in \mathcal G\) and \(b\in \mathcal B^\circ\),
\begin{eqnarray}
f_{g\cdot b}=f_b.
\end{eqnarray}
Moreover, there exists a measurable predictor map \(u\mapsto f_u\) on \(U\) such that
\begin{eqnarray}
f_b=f_{\phi(b)}, \qquad b\in \mathcal B^\circ.
\end{eqnarray}
Thus, the group action changes only the parameter representative. It does not act on the input, output, or predictor. Consequently, the loss, population risk, and empirical risk are constant on each \(\mathcal G\)-orbit and depend on \(b\) only through \(u=\phi(b)\).
\end{assumption*}
The risks can be evaluated directly on the
quotient space, and hence their posterior averages can therefore be computed equivalently in either space.
\begin{proposition}
\label{proposition:1}
Under the parameter-space symmetry assumption, define, for \(u\in U\),
\begin{eqnarray}
R^u(u)
&:=&
\mathbb E_{(X,Y)\sim\mathcal D}
\left[\ell\bigl(f_u(X),Y\bigr)\right],
\\
\widehat R_S^u(u)
&:=&
\frac{1}{n}\sum_{i=1}^{n}
\ell\bigl(f_u(X_i),Y_i\bigr).
\end{eqnarray}
These quotient risks are measurable and satisfy
\begin{eqnarray}
R^b=R^u\circ\phi,
\qquad
\widehat R_S^b=\widehat R_S^u\circ\phi.
\end{eqnarray}
Consequently, for every probability measure \(Q^b\) on \(\mathcal B^\circ\) and its pushforward \(Q^u=\phi_\#Q^b\),
\begin{eqnarray}
\int_{\mathcal B^\circ}R^b(b)\,Q^b(db)
&=&
\int_U R^u(u)\,Q^u(du),
\\
\int_{\mathcal B^\circ}\widehat R_S^b(b)\,Q^b(db)
&=&
\int_U\widehat R_S^u(u)\,Q^u(du).
\end{eqnarray}
\end{proposition}
Instead the parameter-space KL divergence is not invariant since it can include differences between the prior and posterior in how they distribute probability among different parameters belonging to the same symmetry orbit. The quotient map removes this within-orbit information.
\begin{proposition}
\label{proposition:2}
Assume \(Q^b \ll P^b\), and assume additionally that \(\mathcal B^\circ\) and \(\mathcal U\) are standard Borel spaces. Then \(Q^u \ll P^u\). Let \(Q^b(\cdot\mid u)\) and \(P^b(\cdot\mid u)\) be regular conditional distributions of \(Q^b\) and \(P^b\), respectively, given \(u=\phi(b)\). Then
\begin{equation}
\KL(Q^b\|P^b)
=
\KL(Q^u\|P^u)+\Delta_\phi,
\end{equation}
where
\begin{equation}
\label{eq:fiber-gap}
\Delta_\phi
:=
\int_{\mathcal U}
\KL\!\left(Q^b(\cdot\mid u)\,\|\,P^b(\cdot\mid u)\right)
\,Q^u(du)
\ge 0.
\end{equation}
Consequently,
\begin{equation}
\KL(Q^u\|P^u)\le\KL(Q^b\|P^b).
\end{equation}
\end{proposition}
This extra contribution, Eq.~\eqref{eq:fiber-gap}, is nonnegative, so passing to the quotient leaves the two risk terms unchanged and cannot increase the complexity term. We can therefore apply the same PAC--Bayesian inequality on the quotient space.

\begin{theorem}
\label{theorem:1}
Suppose Propositions~\ref{proposition:1} and~\ref{proposition:2} holds. In particular, that \(P^b\) is independent of \(S\), that \(Q^b\ll P^b\), and that the parameter-space symmetry assumption holds. Let \(P^u=\phi_\#P^b\) and \(Q^u=\phi_\#Q^b\). Consider a PAC--Bayesian inequality whose certificate is nondecreasing in the KL divergence. Applied on \(U\), it yields, with probability at least \(1-\delta\) over the draw of the certification sample \(S\),
\begin{eqnarray}
\mathbb E_{u\sim Q^u}[R^u]
\leq
\mathbb E_{u\sim Q^u}[\widehat R_S^u]
+
\operatorname{Comp}_{n,\delta}
\!\left(\KL(Q^u\Vert P^u)\right),
\end{eqnarray}
where \(\operatorname{Comp}_{n,\delta}\) is nondecreasing. Moreover,
\begin{eqnarray}
\begin{split}
\mathbb E_{u\sim Q^u}[R^u]
&=
\mathbb E_{b\sim Q^b}[R^b],
\\
\mathbb E_{u\sim Q^u}[\widehat R_S^u]
&=
\mathbb E_{b\sim Q^b}[\widehat R_S^b],
\end{split}
\end{eqnarray}
\begin{equation}
    \KL(Q^u\Vert P^u) = \KL(Q^b\Vert P^b)-\Delta_{\phi}.
    \label{eq:KL_with_gap}
\end{equation}
where, \(\Delta_\phi\) is defined in Eq.~\eqref{eq:fiber-gap}. Consequently, the quotient certificate is no larger than the certificate obtained by applying the same PAC--Bayesian inequality on \(\mathcal B^\circ\) with \(P^b\) and \(Q^b\).
\end{theorem}

This theorem shows that, for a fixed parameter-space prior and posterior, passing to the quotient preserves the empirical Gibbs risk and cannot increase the KL complexity term. Thus, for any PAC--Bayesian bound whose complexity term is nondecreasing in the KL divergence, quotienting yields a certificate no larger than the corresponding parameter-space certificate. Having established this general PAC--Bayesian advantage of removing redundant parameterizations, we now focus on the quotient space itself. The remaining question is how to select a sample-independent prior on the quotient that is well aligned with the learned quotient posterior. In the next section, we introduce the implicit-bias prior and derive a prior-improvement criterion that provides a particular and sufficient comparison with a neutral reference prior. We evaluate this criterion using numerical experiments in Hadamard and Query-Key attention settings.

\section{Implicit-bias priors}
The quotient result in Theorem~\ref{theorem:1} does not tell which prior is most appropriate. The missing principle comes from our previous geometric analysis of symmetry breaking in redundant stochastic parametrizations~\cite{aladrah2026understanding}. There, we showed that the symmetry of a parameterization determines a geometry-dependent correction, which can be used to construct an implicit-bias prior. It identifies, in closed form, a preference that emerges from the optimization dynamics despite not being imposed as part of the training objective. 

\begin{theorem}[Symmetry-breaking induced implicit bias
\cite{aladrah2026understanding}]
\label{theorem:2}
Let \(\mathcal B^\circ\) be a smooth Riemannian manifold with metric \(g\), and let an \(m\)-dimensional Lie group \(\mathcal G\) act smoothly, freely, and properly on \(\mathcal B^\circ\) by predictor-preserving transformations. Let \(\chi:\mathcal B^\circ\to\mathbb R^m\) be a smooth symmetry-breaking map such that \(0\) is a regular value and \(S_\chi:=\chi^{-1}(0)\) is a transverse slice intersecting each orbit once in the region under consideration. Define
\begin{equation}
\bigl(G_\chi(b)\bigr)_{ij}
:=
\left\langle \nabla\chi_i(b),\nabla\chi_j(b)\right\rangle_g .
\end{equation}
The stationary density induced on \(S_\chi\) is proportional to
\begin{equation}
\exp\!\left[-\frac{\beta}{\sigma^2}L(b)\right]
\det G_\chi(b)^{-1/2},
\end{equation}
with respect to the induced surface measure. Equivalently, the
symmetry-reduced dynamics are associated with the effective loss
\begin{equation}
L_{\mathrm{eff}}(b)
=
L(b)
+
\frac{\sigma^2}{2\beta}\log\det G_\chi(b).
\end{equation}
\end{theorem}
This theorem identifies the geometric factor \(\det G_\chi^{-1/2}\) that represents the implicit bias. Let \(s_\chi:U\to S_\chi\) denote the section induced by the slice, and define the corresponding implicit-bias loss
\begin{equation}
L_\chi(u):=
\frac12\log\det G_\chi\bigl(s_\chi(u)\bigr).
\end{equation}
We use this factor to construct a PAC--Bayesian prior. Given a sample-independent reference prior \(P_0^u\) such that
\begin{equation}
0<Z_\chi:=
\int_U e^{-L_\chi(u)}P_0^u(du)<\infty,
\end{equation}
define
\begin{equation}
\label{eq:ib_prior}
P_{\mathrm{IB}}^u(du)
:=
\frac{e^{-L_\chi(u)}}{Z_\chi}P_0^u(du).
\end{equation}
The remaining question is whether this geometric reweighting actually improves the PAC--Bayesian certificate. The following proposition is the second main result of our analysis. It gives an exact criterion for when the implicit-bias prior has a smaller KL divergence from a given quotient posterior than the reference prior.
\begin{proposition}[Prior-improvement criterion]
\label{proposition:3}
Let \(P_0^u\) be a sample-independent reference prior on \(U\), and let \(P_{\mathrm{IB}}^u\) be the implicit-bias prior defined in Eq.~\eqref{eq:ib_prior}. Let \(Q^u\ll P_0^u\) be a quotient posterior such that \(\mathbb E_{u\sim Q^u}[L_\chi]<\infty\). Define the prior-improvement criterion
\begin{equation}
\Delta\KL(Q^u\Vert P_{\mathrm{IB}}^u)
:=
\KL(Q^u\Vert P_{\mathrm{IB}}^u)
-
\KL(Q^u\Vert P_0^u).
\end{equation}
Then
\begin{equation}
\Delta\KL(Q^u\Vert P_{\mathrm{IB}}^u)
=
\mathbb E_{u\sim Q^u}[L_\chi]+\log Z_\chi.
\end{equation}
Moreover, consider any PAC--Bayesian certificate on \(U\) whose complexity term is nondecreasing in the KL divergence. For the same quotient posterior \(Q^u\), replacing \(P_0^u\) by \(P_{\mathrm{IB}}^u\) leaves the empirical Gibbs risk unchanged. Therefore, whenever
\begin{equation}
\Delta\KL (Q^u\Vert P_{\mathrm{IB}}^u) \leq 0,
\end{equation}
the certificate obtained with \(P_{\mathrm{IB}}^u\) is no larger than the certificate obtained with \(P_0^u\).
\end{proposition}

\subsection{Case study: Hadamard parametrization}
Consider the Hadamard parametrization of the weights \(w\), wuch that \(w=a\odot c\), on the regular parameter space \(\mathcal B^\circ=(\mathbb R^\times)^{2d}\). The group \(\mathcal G=(\mathbb R^\times)^d\) acts by coordinatewise rescaling,
\begin{eqnarray}
\lambda\cdot(a,c)
:=
\left(
\lambda\odot a,\,
\lambda^{-1}\odot c
\right),
\qquad
\lambda\in\mathcal G.
\end{eqnarray}
This action preserves the weights \(w\), since \((\lambda\odot a)\odot(\lambda^{-1}\odot c)=a\odot c\). Hence the quotient is identified with \(U=(\mathbb R^\times)^d\), and the quotient map is \(\phi(a,c)=a\odot c=w\). Following the symmetry-breaking construction of~\cite{aladrah2026understanding}, for the canonical slice
\begin{eqnarray}
\chi_i(a,c):=\frac12(a_i^2-c_i^2)=0,
\qquad i=1,\ldots,d,
\end{eqnarray}
To choose one representative on each orbit, we impose the branch convention
\(a_i>0\). We next compute the geometric factor generated by this slice. Under the Euclidean metric on \(\mathcal B^\circ\), the gradient of the \(i\)-th constraint has nonzero components \(a_i\) and \(-c_i\). Consequently, the constraint Gram matrix is diagonal:
\begin{eqnarray}
\bigl(G_\chi(a,c)\bigr)_{ij}
=
\delta_{ij}(a_i^2+c_i^2).
\end{eqnarray}
Evaluating it on the balanced section we have
\begin{eqnarray}
\det G_\chi\bigl(s_\chi(w)\bigr)
&=&
\prod_{i=1}^{d}2|w_i|.
\end{eqnarray}
The resulting implicit-bias loss is therefore
\begin{eqnarray}
L_\chi(w)
&=&
\frac{1}{2}
\sum_{i=1}^{d}
\log\bigl(2|w_i|\bigr).
\label{eq:ib_loss}
\end{eqnarray}
Let \(P_0^w\) be a reference prior on \(U=(\mathbb R^\times)^d\), fixed independently of the certification sample, and assume that the following normalizing constant is finite:
\begin{eqnarray}
Z_{\mathrm{IB}}
:=
\int_U
\prod_{i=1}^{d}|w_i|^{-1/2}\,
P_0^w(dw).
\end{eqnarray}
The exact Hadamard implicit-bias prior is
\begin{eqnarray}
P_{\mathrm{IB}}^w(dw)
&:=&
\frac{1}{Z_{\mathrm{IB}}}
\prod_{i=1}^{d}|w_i|^{-1/2}\,
P_0^w(dw).
\end{eqnarray}
Here, the constant factor \(2^{-d/2}\) from
\(\exp[-L_\chi(w)]\) has been absorbed into
\(Z_{\mathrm{IB}}\). Thus, the symmetry-induced logarithmic loss in Eq.~\eqref{eq:ib_loss} enters the PAC--Bayesian analysis as a modified prior on the weights \(w\).

\begin{table*}[t]
\centering
\caption{PAC--Bayesian certificates at the reference scale \(s_0=20\). Entries are mean \(\pm\) sample standard deviation over 40 optimizer seeds.\\}
\label{tab:reference-certificates}
\small
\begin{tabular}{lcccc}
\toprule
Experiment & Certification prior &
\(\mathbb E_{Q}[\widehat R_S]\) &
\(\KL(Q\Vert P)\) &
PAC--Bayes bound \\
\midrule
Fourier Naive dense & \(P_0\) &
\(0.219530\pm0.000235\) &
\(626.468306\pm0.860165\) &
\(0.346158\pm0.000243\) \\
\multirow{2}{*}{Hadamard parameterization} & \(P_0\) &
\(0.005340\pm0.000541\) &
\(585.574684\pm5.616454\) &
\(0.127863\pm0.000932\) \\
& \(P_{\mathrm{IB}}\) &
\(0.005340\pm0.000541\) &
\(347.300316\pm4.197013\) &
\(0.100498\pm0.000935\) \\
\midrule
\multirow{2}{*}{Query--Key} & \(P_{0}\)&
\(0.061372\pm0.005501\) &
\(2288.319332\pm14.972336\) &
\(0.301333\pm0.004755\) \\
 & \(P_{\mathrm{IB}}\) &
\(0.061372\pm0.005501\) &
\(2263.264058\pm14.957115\) &
\(0.300024\pm0.004752\) \\
\bottomrule
\end{tabular}
\end{table*}

\section{Experiments}
\label{sec:experiments}

We evaluate the prior-improvement criterion from Proposition~\ref{proposition:3} in two controlled settings: sparse Fourier regression with a Hadamard parameterization and a single-head Query-Key attention teacher--student model. In both experiments, training uses only the ordinary mean-squared-error objective. No implicit-bias term is added during training. After a burn-in period (post-burn-in), we record SGD iterates and use them to fit an isotropic Gaussian posterior. For each fitted posterior, the posterior distribution and empirical Gibbs risk are held fixed while the neutral and implicit-bias priors are compared. The complete certification protocol is given in Appendix~\ref{app:experimental-details}.

For a fixed fitted posterior \(Q\) and neutral-prior scale \(s_0\), we measure
\begin{align}
\Delta\KL(Q\Vert P^{(s_0)}_{\mathrm{IB}})
&:=
\KL\!\left(Q\middle\Vert
P_{\mathrm{IB},\varepsilon}^{(s_0)}\right)
-
\KL\!\left(Q\middle\Vert P_0^{(s_0)}\right)\\
&=
\mathbb E_{Q}[L_\varepsilon]+\log Z_\varepsilon(s_0).
\label{eq:experimental-delta}
\end{align}
Consequently, \(\Delta\KL(Q\Vert P^{(s_0)}_{\mathrm{IB}})<0\) is exactly the condition under which the implicit-bias prior decreases the prior-dependent KL term.

\subsection{Sparse Fourier regression}

The data-generating predictor is a sparse Fourier series in a cosine basis of dimension \(d=64\). Its coefficient vector \(w^\star\in\mathbb R^{64}\) has three nonzero entries, \(w^\star_3=1.2\), \(w^\star_9=-0.9\) and \(w^\star_{17}=0.7\). For inputs \(t_i\) drawn uniformly from \([0,1]\), let \(\Phi(t_i)\in\mathbb R^{64}\) denote the corresponding cosine-feature vector. We generate \(n_{\mathrm{train}}=32\) training data as
\begin{eqnarray}
y_i
&=&
\Phi(t_i)^\top w^\star+\xi_i,
\end{eqnarray}
where \(\xi_i\) is Gaussian noise with standard deviation equal to \(5\%\) of the largest absolute noiseless training response. An independent certification sample of size \(n_{\mathrm{cert}}=20{,}000\) is drawn from the same distribution.

We compare two linear models. The naive dense model uses \(w\in\mathbb R^{64}\) and predicts
\begin{eqnarray}
f_w(t)
&=&
\Phi(t)^\top w.
\end{eqnarray}
The Hadamard model uses \((a,c)\in\mathbb R^{64}\times\mathbb R^{64}\) and predicts
\begin{eqnarray}
f_{a,c}(t)
&=&
\Phi(t)^\top(a\odot c).
\end{eqnarray}
Both models are trained for \(70{,}000\) SGD steps. We use batch size \(16\), learning rate \(10^{-3}\), zero momentum, and zero weight decay. No implicit-bias term is added to the training objective. For each optimizer seed, the weight \(w\) or \(w=a\odot c\), is retained every \(200\) updates from update \(60{,}000\) through update \(70{,}000\). This gives \(T=51\) post-burn-in trajectory snapshots.

For the naive model, the retained weight is \(w^{(t)}\). For the Hadamard model, it is \(w^{(t)}=a^{(t)}\odot b^{(t)}\), rather than the redundant parameter pair \((a^{(t)},b^{(t)})\). These retained SGD states are used to fit the posterior. They are not treated as independent posterior samples.

The neutral reference prior on the quotient space is the isotropic Gaussian \(P^{w,(s_0)}_0=\mathcal N(0,s_0^2I_{64})\). The implicit-bias prior is obtained by tilting this reference prior with the implicit-bias loss. The corresponding implicit-bias prior and its normalizing constant are constructed after training. Their numerical evaluation is described in Appendix~\ref{app:fourier-certification}.

\subsection{Query-Key attention}

We consider a single-head attention predictor with sequence length \(L=8\),
embedding dimension \(d_e=16\), and head dimension \(d_h=8\). Its parameters
are
\[
\theta=(W_Q,W_K,W_V),
\qquad
W_Q,W_K,W_V\in\mathbb R^{d_e\times d_h}.
\]
For an input sequence \(X\in\mathbb R^{L\times d_e}\), the predictor is
\begin{align}
f_\theta(X)
:=
\operatorname{softmax}\!\left(
\frac{(XW_Q)(XW_K)^{\mathsf T}}{\sqrt{d_h}}
\right)XW_V,
\end{align}
where the softmax is applied row-wise. A frozen teacher of the same form
generates the labels
\begin{align}
Y=f_{\theta^\star}(X)+E,
\end{align}
where the entries of \(X\) are independent standard Gaussian variables and the entries of \(E\) are independent centered Gaussian variables with standard deviation \(0.02\).

Posterior fitting begins after \(60\%\) of training. Every \(200\) updates from update \(60{,}000\) through update \(100{,}000\), the Query and Key matrices are mapped to the implicit-bias loss, \(L_{\chi,\varepsilon}\), and the vector \(z^{(t)} = \operatorname{vec} \left( \bar W_Q^{(t)},\bar W_K^{(t)},W_V^{(t)} \right) \in\mathbb R^{384}\). is retained. This gives \(T=201\) trajectory snapshots. An isotropic Gaussian \(Q^z\) is fitted to these vectors. The snapshots are observations from one correlated SGD trajectory and are used only to estimate the mean and scalar variance of \(Q^z\). They are not treated as independent posterior samples as illustrated in~Appendix~\ref{app:query-key-certification}.

For a fixed fitted posterior \(Q^z\), the neutral reference prior is \(P_0^{z,(s_0)}=\mathcal N(0,s_0^2I_{384})\). The implicit-bias prior is obtained by tilting this Gaussian by \(\exp[-L_{\chi,\varepsilon}]\). Both certificates use the same \(Q^z\) and the same empirical Gibbs risk. Their difference therefore isolates the prior-dependent KL term. 

We train using plain SGD. Each update uses a newly generated batch of \(64\) teacher examples. We use learning rate \(10^{-3}\), zero momentum, zero weight decay, and \(100{,}000\) steps. Similarly to the Fourier experiment, no implicit-bias term is added to the training objective.

\subsection{Results}
\label{sec:joint-results}
Table~\ref{tab:reference-certificates} reports the results at the reference scale \(s_0=20\). For the fitted Fourier Hadamard posterior, replacing the neutral prior by the regularized implicit-bias prior reduces the mean KL divergence from \(585.574684\) to \(347.300316\), a reduction of \(40.69\%\). The corresponding mean certificate decreases from \(0.127863\) to \(0.100498\), a reduction of \(21.40\%\).
\begin{table}[!htbp]
\centering
\caption{Mean \(\pm\) sample standard deviation of the prior-improvement quantity, \(\Delta\KL (Q^u\Vert P^u_{\mathrm{IB}})\), over \(40\) optimizer seeds. Negative values indicate that the implicit-bias prior has the smaller KL divergence.\\}
\label{tab:prior-scale-sensitivity}
\renewcommand{\arraystretch}{1.08}
\begin{tabular*}{\columnwidth}{@{\extracolsep{\fill}}c cc}
\toprule
\(s_0\) &
Fourier &
Query--Key \\
\midrule
0.25 & \(-108.351\pm5.144\) & \(9.999\pm0.300\) \\
0.50 & \(-129.442\pm5.142\) & \(4.455\pm0.299\) \\
0.75 & \(-141.936\pm5.141\) & \(1.212\pm0.300\) \\
1.00 & \(-150.851\pm5.140\) & \(-1.089\pm0.299\) \\
1.25 & \(-157.797\pm5.137\) & \(-2.874\pm0.300\) \\
1.50 & \(-163.486\pm5.138\) & \(-4.332\pm0.299\) \\
2.00 & \(-172.487\pm5.138\) & \(-6.634\pm0.299\) \\
5.00 & \(-201.323\pm5.139\) & \(-13.965\pm0.299\) \\
7.00 & \(-211.962\pm5.138\) & \(-16.656\pm0.299\) \\
10.00 & \(-223.258\pm5.139\) & \(-19.510\pm0.299\) \\
20.00 & \(-245.266\pm5.136\) & \(-25.055\pm0.300\) \\
\bottomrule
\end{tabular*}
\end{table}
These two rows use the same posterior and empirical Gibbs risk, so the difference is caused entirely by the prior-dependent KL term.

For the Query-Key implicit-bias prior at \(s_0=20\), the mean KL divergence decreases from \(2288.319332\) to \(2263.264058\), a reduction of \(1.09\%\). The mean certificate decreases from \(0.301333\) to \(0.300024\), a reduction of \(0.43\%\). Thus, the prior improvement is substantially smaller in the Query-Key experiment than in the Fourier experiment comparison. This is not a surprising result. The smaller relative improvement in can be understood from the number of symmetry directions included in the implicit-bias term, \(d_h\). In the Fourier-Hadamard experiment, the correction acts across all \(d=64\) independently. While in the Query-Key experiment, the evaluated correction acts only along the \(d_h=8\) head-wise symmetry directions, while the neutral-prior KL is computed for the full \(384\)-dimensional vector containing \(W_Q\), \(W_K\), and \(W_V\). This explain why the total KL reduction is much smaller as the implicit-bias loss modifies only a restricted part of the complexity term.

Table~\ref{tab:prior-scale-sensitivity} shows that the mean \(\Delta\KL(Q^u\Vert P^u_{\mathrm{IB}})\) is negative at all 11 tested scales in the Fourier-Hadamard experiment. In the Query-Key experiment, its mean is positive at \(s_0\in\{0.25,0.5,0.75\}\) and negative at the tested scales \(s_0\in\{1,1.25,1.5,2,5,7,10,20\}\). Therefore, within the tested grid, the implicit-bias prior does not improve the Query-Key comparison at the three smallest scales.
\section{Conclusion}

This work separates predictor-level complexity from representational complexity in PAC--Bayesian analysis. For a predictor-preserving parameter symmetry, pushing a parameter-space prior and posterior through the quotient map leaves their posterior-averaged empirical and population risks unchanged. The KL divergence cannot increase, and its exact reduction is the conditional posterior--prior divergence within symmetry orbits.

The geometric construction addresses the separate problem of selecting a prior on the quotient space. Because the parameterization possesses a predictor-preserving symmetry, the symmetry-breaking geometry induces an additional preference in the SGD dynamics, implicitly, even though no explicit regularizer is added to the training objective. This preference is represented by a volume correction, which defines an implicit-bias prior through the tilting of a normalizable reference prior. We show that this symmetry-induced preference can improve the PAC--Bayes certificate without any intervention in the training procedure: when the learned posterior is sufficiently aligned with the implicit-bias prior, the prior-dependent KL term decreases and the certificate becomes tighter, precisely when \(\Delta\KL(Q\Vert P_{\mathrm{IB}})<0\). The improvement is therefore implicit and conditional, arising from the interaction between the symmetry-induced geometry and the posterior produced by SGD.

The experiments evaluate the posterior-dependent prior comparison rather than the quotienting inequality itself. In the Fourier-Hadamard experiment, the regularized implicit-bias prior has a smaller mean KL divergence than the neutral prior at every tested scale. In the Query-Key experiment, the mean KL difference is positive at the three tested scales below \(1\) and negative at every tested scale from \(1\) to \(20\). The result therefore depends on both the learned posterior and the reference-prior scale.

This perspective suggests a connection with inverse learning of implicit bias. Given a family of predictor-preserving parameterizations, one could seek a design whose induced implicit-bias prior is well matched to a specified class of posteriors. Such a procedure must choose the parametrization and prior family independently of the certification sample. Developing sample-independent procedures for this inverse-learning problem and constructing Bayesian quotient posteriors are natural directions for future work.

\section*{Data and Code Availability}

The data and code used for the numerical experiments are available from the authors upon reasonable request.
\printbibliography

@article{aladrah2026understanding,
  title={Understanding and inverse design of implicit bias in stochastic learning: a geometric perspective},
  author={Aladrah, Nicola and Ballarin, Emanuele and Biagetti, Matteo and Ansuini, Alessio and d’Onofrio, Alberto and Anselmi, Fabio},
  journal={arXiv preprint arXiv:2601.06597},
  year={2026}
}

@article{ziyin2025parameter,
  title={Parameter symmetry potentially unifies deep learning theory},
  author={Ziyin, Liu and Xu, Yizhou and Poggio, Tomaso and Chuang, Isaac},
  journal={arXiv preprint arXiv:2502.05300},
  year={2025}
}

@inproceedings{mcallester1998some,
  title={Some pac-bayesian theorems},
  author={McAllester, David A},
  booktitle={Proceedings of the eleventh annual conference on Computational learning theory},
  pages={230--234},
  year={1998}
}

@article{mcallester2013pac,
  title={A PAC-Bayesian tutorial with a dropout bound},
  author={McAllester, David},
  journal={arXiv preprint arXiv:1307.2118},
  year={2013}
}

@article{beck2025symmetries,
  title={Symmetries in PAC-Bayesian Learning},
  author={Beck, Armin and Ochs, Peter},
  journal={arXiv preprint arXiv:2510.17303},
  year={2025}
}

@article{alquier2016,
  author  = {Pierre Alquier and James Ridgway and Nicolas Chopin},
  title   = {On the properties of variational approximations of Gibbs posteriors},
  journal = {Journal of Machine Learning Research},
  year    = {2016},
  volume  = {17},
  number  = {236},
  pages   = {1--41},
  url     = {http://jmlr.org/papers/v17/15-290.html}
}

@inproceedings{shalaeva2020improved,
  title={Improved PAC-Bayesian bounds for linear regression},
  author={Shalaeva, Vera and Esfahani, Alireza Fakhrizadeh and Germain, Pascal and Petreczky, Mihaly},
  booktitle={Proceedings of the AAAI Conference on Artificial Intelligence},
  volume={34},
  number={04},
  pages={5660--5667},
  year={2020}
}

@article{seeger2002pac,
  title={PAC-Bayesian generalisation error bounds for Gaussian process classification},
  author={Seeger, Matthias},
  journal={Journal of machine learning research},
  volume={3},
  number={Oct},
  pages={233--269},
  year={2002}
}

@article{lyle2020benefits,
  title={On the benefits of invariance in neural networks},
  author={Lyle, Clare and van der Wilk, Mark and Kwiatkowska, Marta and Gal, Yarin and Bloem-Reddy, Benjamin},
  journal={arXiv preprint arXiv:2005.00178},
  year={2020}
}

@article{catoni2007pac,
  title={PAC-Bayesian supervised classification: the thermodynamics of statistical learning},
  author={Catoni, Olivier},
  journal={arXiv preprint arXiv:0712.0248},
  year={2007}
}

@article{zhao2025symmetry,
  title={Symmetry in neural network parameter spaces},
  author={Zhao, Bo and Walters, Robin and Yu, Rose},
  journal={arXiv preprint arXiv:2506.13018},
  year={2025}
}

@inproceedings{simsek2021geometry,
  title={Geometry of the loss landscape in overparameterized neural networks: Symmetries and invariances},
  author={Simsek, Berfin and Ged, Fran{\c{c}}ois and Jacot, Arthur and Spadaro, Francesco and Hongler, Cl{\'e}ment and Gerstner, Wulfram and Brea, Johanni},
  booktitle={International Conference on Machine Learning},
  pages={9722--9732},
  year={2021},
  organization={PMLR}
}

@inproceedings{zhao2024improving,
  title={Improving convergence and generalization using parameter symmetries},
  author={Zhao, Bo and Gower, Robert M and Walters, Robin and Yu, Rose},
  booktitle={International Conference on Learning Representations},
  volume={2024},
  pages={55008--55035},
  year={2024}
}

@article{godfrey2022symmetries,
  title={On the symmetries of deep learning models and their internal representations},
  author={Godfrey, Charles and Brown, Davis and Emerson, Tegan and Kvinge, Henry},
  journal={Advances in Neural Information Processing Systems},
  volume={35},
  pages={11893--11905},
  year={2022}
}

@article{mcallester2003pac,
  title={PAC-Bayesian stochastic model selection},
  author={McAllester, David A},
  journal={Machine Learning},
  volume={51},
  number={1},
  pages={5--21},
  year={2003},
  publisher={Springer}
}
\appendix

\renewcommand{\theequation}{\thesection.\arabic{equation}}
\counterwithin{equation}{section}
\section{Proofs}
\label{app:quotient-proofs}
This appendix proves the quotient-risk identities, the KL decomposition along symmetry orbits, and the prior-comparison identity used in the main text. The proofs are presented in the order of the corresponding propositions and theorem.
\subsection{Proposition~\ref{proposition:1}}
\label{app:proof_prop_1}
By the parameter-space symmetry assumption, there exists a measurable predictor family \(u \mapsto f^u\) such that \(f^b = f^{\phi(b)}\) for every \(b \in \mathcal{B}^{\circ}\). Hence, for every \(b \in \mathcal{B}^{\circ}\),
\begin{align}
R^b(b)
&= \mathbb{E}_{(X,Y)\sim\mathcal{D}}
   \left[\ell\!\left(f^{\phi(b)}(X),Y\right)\right]
 = R^u(\phi(b)), \\
\widehat R_S^b(b)
&= \frac{1}{n}\sum_{i=1}^n
   \ell\!\left(f^{\phi(b)}(X_i),Y_i\right)
 = \widehat R_S^u(\phi(b)).
\end{align}
Let \(Q^u=\phi_{\#}Q^b\). For every integrable measurable \(h:U \to\mathbb{R}\), the defining property of the pushforward gives
\begin{equation}
\int_{U} h(u)\,Q^u(du)
=
\int_{\mathcal{B}^{\circ}} h(\phi(b))\,Q^b(db).
\end{equation}
Applying this identity to \(h=R^u\) and \(h=\widehat R_S^u\), and using the identities above, proves the two posterior-risk equalities.

\subsection{Proposition~\ref{proposition:2}}
\label{app:proof_prop_2}
Let \(A\subseteq U\) be measurable and suppose that \(P^u(A)=0\). Since \(P^u=\phi_{\#}P^b\), \(P^b\!\left(\phi^{-1}(A)\right)=P^u(A)=0\). Because \(Q^b\ll P^b\), \(Q^u(A) = Q^b\!\left(\phi^{-1}(A)\right) = 0\). Hence \(Q^u\ll P^u\).

Since \(\mathcal B^\circ\) and \(U\) are standard Borel
spaces, and given \(u=\phi(b)\) the disintegrations,
\begin{align}
Q^b(db)=Q^b(db\mid u)\,Q^u(du),
\nonumber\\
P^b(db)=P^b(db\mid u)\,P^u(du).
\end{align}
Moreover, \(Q^b\ll P^b\) and \(Q^u\ll P^u\) imply that \(Q^b(\cdot\mid u)\ll P^b(\cdot\mid u)\) for \(Q^u\)-almost every \(u\).

The Radon--Nikodym derivative therefore factorizes
\(Q^b\)-almost surely as
\begin{equation}
\frac{dQ^b}{dP^b}(b)
=
\frac{dQ^u}{dP^u}(u)
\frac{dQ^b(\cdot\mid u)}{dP^b(\cdot\mid u)}(b).
\end{equation}
Taking logarithms and integrating with respect to \(Q^b\), followed by
the disintegration of \(Q^b\), gives the relative-entropy chain rule
\begin{align}
\KL(Q^b\|P^b)
&=
\KL(Q^u\|P^u)
\nonumber\\
&+
\int_{U}
\KL\!\left(
Q^b(\cdot\mid u)\,\middle\|\,P^b(\cdot\mid u)
\right)
Q^u(du).
\end{align}
The second term is nonnegative, which proves the claimed inequality.

\subsection{Theorem~\ref{theorem:1}}
\label{app:proof_theorem_1}
Since \(P^b\) is independent of the certification sample \(S\) and \(\phi\) is fixed independently of \(S\), its pushforward \(P^u=\phi_{\#}P^b\) is also independent of \(S\). Therefore \(P^u\) is an admissible PAC--Bayes prior on \(U\). Applying the selected PAC--Bayes inequality on \(U\) gives, with probability at least \(1-\delta\),
\begin{equation}
\mathbb E_{u\sim Q^u}[R^u]
\le
\mathbb E_{u\sim Q^u}[\widehat R_S^u]
+
\mathrm{Comp}_{n,\delta}\!\left(\KL(Q^u\|P^u)\right).
\end{equation}
By Proposition~\ref{proposition:1},
\begin{align}
\mathbb E_{u\sim Q^u}[R^u]&=\mathbb E_{b\sim Q^b}[R^b],
\nonumber\\
\mathbb E_{u\sim Q^u}[\widehat R_S^u]&=\mathbb E_{b\sim Q^b}[\widehat R_S^b].
\end{align}
By Proposition~\ref{proposition:2},
\begin{equation}
\KL(Q^u\|P^u)\le\KL(Q^b\|P^b).
\end{equation}
Since \(\mathrm{Comp}_{n,\delta}\) is nondecreasing,
\begin{equation}
\mathrm{Comp}_{n,\delta}\!\left(\KL(Q^u\|P^u)\right)
\le
\mathrm{Comp}_{n,\delta}\!\left(\KL(Q^b\|P^b)\right).
\end{equation}
Combining these identities and inequalities yields
\begin{align}
\begin{aligned}
&\mathbb E_{u\sim Q^u}[\widehat R_S^u]
+
\mathrm{Comp}_{n,\delta}\!\left(\KL(Q^u\|P^u)\right) \\
&\qquad\le
\mathbb E_{b\sim Q^b}[\widehat R_S^b]
+
\mathrm{Comp}_{n,\delta}\!\left(\KL(Q^b\|P^b)\right),
\end{aligned}
\end{align}
which proves the theorem.

\subsection{Proposition~\ref{proposition:3}}
\label{app:proof_proposition_3}
By construction,
\begin{equation}
\frac{dP^u_{\mathrm{IB}}}{dP^u_0}(u)
=
\frac{e^{-L_\chi(u)}}{Z_\chi}.
\end{equation}
Since \(L_\chi\) is finite \(P^u_0\)-almost everywhere and \(0<Z_\chi<\infty\), this density is strictly positive \(P^u_0\)-almost everywhere. Hence \(Q^u\ll P^u_0\) implies \(Q^u\ll P^u_{\mathrm{IB}}\). Moreover,
\begin{equation}
\frac{dP^u_0}{dP^u_{\mathrm{IB}}}(u)
=
Z_\chi e^{L_\chi(u)}.
\end{equation}
Therefore,
\begin{equation}
\frac{dQ^u}{dP^u_{\mathrm{IB}}}(u)
=
\frac{dQ^u}{dP^u_0}(u)
\frac{dP^u_0}{dP^u_{\mathrm{IB}}}(u),
\end{equation}
and hence
\begin{equation}
\log\frac{dQ^u}{dP^u_{\mathrm{IB}}}(u)
=
\log\frac{dQ^u}{dP^u_0}(u)
+
L_\chi(u)
+
\log Z_\chi.
\end{equation}
The assumptions \(\KL(Q^u\|P^u_0)<\infty\) and \(\mathbb E_{u\sim Q^u}[|L_\chi|]<\infty\) permit integration with respect to \(Q^u\). Thus,
\begin{equation}
\KL(Q^u\|P^u_{\mathrm{IB}})
=
\KL(Q^u\|P^u_0)
+
\mathbb E_{u\sim Q^u}[L_\chi]
+
\log Z_\chi.
\end{equation}
Rearranging gives
\begin{equation}
\Delta\KL(Q^u)
=
\mathbb E_{u\sim Q^u}[L_\chi]+\log Z_\chi.
\end{equation}
For a fixed quotient posterior \(Q^u\), replacing \(P^u_0\) by \(P^u_{\mathrm{IB}}\) does not change either the empirical or the population Gibbs risk. If \(\Delta\KL(Q^u)\le0\), then
\begin{equation}
\KL(Q^u\|P^u_{\mathrm{IB}})
\le
\KL(Q^u\|P^u_0).
\end{equation}
Monotonicity of the PAC--Bayes complexity term then yields the stated comparison of certificate expressions.

\section{Experimental details}
\label{app:experimental-details}

\subsection{Posterior construction and certification protocol}
For retained trajectory vectors \(z^{(1)},\ldots,z^{(T)}\in\mathbb R^p\), we fit
\begin{align}
Q^z
&=
\mathcal N(\bar z,\widehat s_Q^2I_p),\\
\bar z
&=
\frac1T\sum_{t=1}^Tz^{(t)},\\
\widehat s_Q^2
&=
\frac{1}{p(T-1)}
\sum_{t=1}^T\lVert z^{(t)}-\bar z\rVert_2^2.
\end{align}
The fitted posterior is constructed separately for every run. The retained SGD iterates are correlated trajectory observations and are used only to estimate \(\bar z\) and \(\widehat s_Q^2\).

For each tested scale \(s_0\), the neutral prior is \(P_0^{z,(s_0)}=\mathcal N(0,s_0^2I_p)\). Its KL divergence from \(Q^z\) is evaluated analytically. Both experiments use the clipped squared certification loss
\begin{equation}
\ell_M(\widehat y,y)
=
\min\left\{\frac{(\widehat y-y)^2}{M^2},1\right\},
\qquad M=2,
\end{equation}
an independent certification sample of size \(20{,}000\), confidence level \(\delta=0.05\), and \(8{,}192\) draws from the fitted posterior to estimate the empirical Gibbs risk. For a fixed run, changing \(s_0\) changes the two priors but not the fitted posterior distribution or its empirical Gibbs
risk.

\subsection{Fourier--Hadamard implicit-bias loss}
\label{app:fourier-certification}
For both models, the posterior is by the weights \(w\in\mathbb R^{64}\). It is fitted from \(T=51\) SGD states retained every \(200\) updates between updates \(60{,}000\) and \(70{,}000\). For the Hadamard model, the retained state is \(w=a\odot c\), rather than the redundant pair \((a,c)\).

The regularized implicit-bias loss for \(\varepsilon=10^{-3}\) is
\begin{equation}
L_{\chi,\varepsilon}(w)
=
\frac12\sum_{i=1}^{64}\log(|w_i|+\varepsilon).
\end{equation}
The corresponding implicit-bias prior is
\begin{equation}
P_{\mathrm{IB},\varepsilon}^{w,(s_0)}(dw)
=
\frac{e^{-L_{\chi,\varepsilon}(w)}}
{Z_{\mathrm{IB},\varepsilon}(s_0)}
P_0^{w,(s_0)}(dw).
\end{equation}
Because the Gaussian reference prior and the implicit-bias loss factorize over weights, the normalizer reduces to a one-dimensional integral. This integral is approximated by trapezoidal quadrature on \([-160,160]\) using \(800{,}001\) equally spaced points. The expectation \(\mathbb E_{Q}[L_{\chi,\varepsilon}]\) is estimated with \(8{,}192\) posterior draws.

\subsection{Query--Key implicit-bias loss}
\label{app:query-key-certification}

For the query--key model, the weights are \(z=\operatorname{vec}(W_Q,W_K,W_V)\in\mathbb R^{384}\). The posterior is fitted from \(T=201\) balanced trajectory vectors retained every \(200\) updates between updates \(60{,}000\) and \(100{,}000\). Each vector \(z\in\mathbb R^{384}\) contains the balanced Query and Key matrices together with \(W_V\).

To define the balanced vector, let \(q_j\) and \(k_j\) denote the \(j\)-th columns of \(W_Q\) and \(W_K\), respectively. For nonzero columns, define
\begin{equation}
\bar q_j
=
q_j
\sqrt{\frac{\|k_j\|_2}{\|q_j\|_2}},
\qquad
\bar k_j
=
k_j
\sqrt{\frac{\|q_j\|_2}{\|k_j\|_2}},
\qquad
j=1,\ldots,d_h.
\end{equation}
This transformation preserves the Query--Key predictor and satisfies the balanced condition~\cite{aladrah2026understanding}, 
\begin{equation}
\|\bar q_j\|_2=\|\bar k_j\|_2.
\end{equation}
We then define
\begin{equation}
\operatorname{bal}(z)
=
\operatorname{vec}(\bar W_Q,\bar W_K,W_V),
\end{equation}
where \(\bar W_Q\) and \(\bar W_K\) have columns \(\bar q_j\) and \(\bar k_j\). The corresponding regularized implicit-bias loss, for \(\varepsilon=10^{-3}\) is
\begin{equation}
L_{\chi,\varepsilon}(\operatorname{bal}(z))
=
\frac12\sum_{j=1}^{d_h}
\log\!\left(
\|\bar q_j\|_2^2+\|\bar k_j\|_2^2+\varepsilon
\right).
\end{equation}

The expectation of this loss under \(Q^z\) is estimated using \(8{,}192\) posterior draws. The normalizing constant \(Z_{\mathrm{IB}}(s_0)\) is estimated using \(16{,}384\) independent draws from the neutral prior \(P_0^{z,(s_0)}\).
\end{document}